\documentclass[10pt,twocolumn,letterpaper]{article}

\usepackage{cvpr}
\usepackage{times}
\usepackage{epsfig}
\usepackage{graphicx}
\usepackage{amsmath}
\usepackage{amssymb}

\usepackage[pagebackref=true,breaklinks=true,letterpaper=true,colorlinks,bookmarks=false]{hyperref}

 \cvprfinalcopy 

\def\cvprPaperID{5814} 

\ifcvprfinal\fi
\begin{document}

\title{Stacked Spatio-Temporal Graph Convolutional Networks for Action Segmentation}

\author{Pallabi Ghosh\\
University of Maryland\\
College Park\\
{\tt\small pallabig@umd.edu}
\and
Yi Yao\\
SRI International\\
Princeton, NJ\\
{\tt\small yi.yao@sri.com}
\and
Larry Davis\\
University of Maryland\\
College Park\\
{\tt\small lsd@umiacs.umd.edu}
\and
Ajay Divakaran\\
SRI International\\
Princeton, NJ\\
{\tt\small ajay.divakaran@sri.com}
}

\maketitle

\begin{abstract}
We propose novel Stacked Spatio-Temporal Graph Convolutional Networks (Stacked-STGCN) for action segmentation, i.e., predicting and localizing a sequence of actions over long videos. We extend the Spatio-Temporal Graph Convolutional Network (STGCN) originally proposed for skeleton-based action recognition to enable nodes with different characteristics (e.g., scene, actor, object, action, etc.), feature descriptors with varied lengths, and arbitrary temporal edge connections to account for large graph deformation commonly associated with complex activities. We further introduce the stacked hourglass architecture to STGCN to leverage the advantages of an encoder-decoder design for improved generalization performance and localization accuracy. We explore various descriptors such as frame-level VGG, segment-level I3D, RCNN-based object, etc. as node descriptors to enable action segmentation based on joint inference over comprehensive contextual information. We show results on CAD120 (which provides pre-computed node features and edge weights for fair performance comparison across algorithms) as well as a more complex real-world activity dataset, Charades. Our Stacked-STGCN in general achieves $4.0\%$ performance improvement over the best reported results in F1 score on CAD120 and $1.3\%$ in mAP on Charades using VGG features.
\end{abstract}

\section{Introduction}

\begin{figure*}
\begin{center}
\fbox{\rule{0pt}{2in} \includegraphics[width=\linewidth]{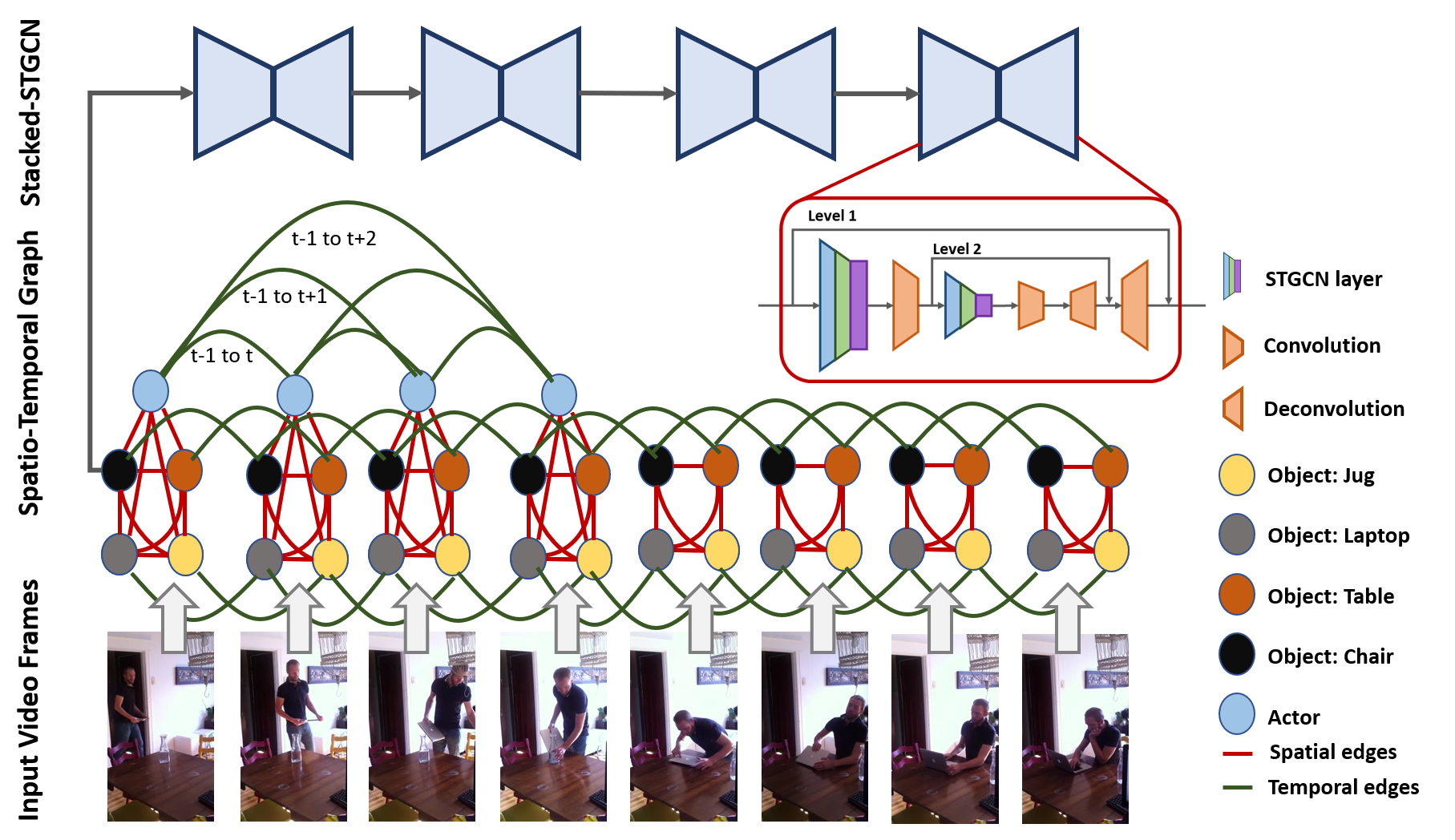}}
\end{center}
   \caption{System overview. The nodes are of various types such as actors, objects, scenes etc. Red lines show spatial connections and green lines temporal connections. Different from the original STGCN based on human skeleton~\cite{stgcn2018aaai}, our graph allows flexible temporal connections that can span multiple time steps, for example the connections among the blue nodes (the actor node). Note that other nodes can have such temporal connections but are not depicted to avoid congested illustration. This spatio-temporal graph is fed into a stack of hourglass STGCN blocks to output a sequence of predicted actions observed in the video.}
\label{fig:system}
\end{figure*}

Inspired by the success of convolutional neural networks (on either grid-like or sequential data), graph neural networks (GNNs) including graph convolutional networks (GCNs) have been developed and 
have demonstrated improvements over a number of machine learning/computer vision tasks such as node classification \cite{kipf2016semi}, community clustering \cite{bruna2017community}, link prediction \cite{schlichtkrull2018modeling}, $3D$ point cloud segmentation~\cite{te2018rgcnn}, etc.

As a special case of GCNs, spatio-temporal graph convolutional networks (STGCN), have been proposed for skeleton-based activity recognition \cite{stgcn2018aaai}. STGCN leverages the spatial connection between the joints of the human body and connects the same joints across time to form a spatio-temporal graph. STGCN has shown performance improvements on Kinetics-skeleton \cite{kay2017kinetics} and NTU RGB+D \cite{shahroudy2016ntu} datasets via exploiting primarily actor poses.  

In addition to actor poses, there frequently exist abundant contextual cues that would help in recognizing an action. Leveraging these contextual cues becomes critical for improving accuracy and robustness of action recognition/segmentation, especially for actions with subtle changes in the actor's movement/pose. A graph is an intuitive data structure to jointly represent various contextual cues (e.g., scene graph, situation recognition, etc.). Therefore, in this paper, we plan to construct a comprehensive spatio-temporal graph (STG) to jointly represent an action along with its associated actors, objects, and other contextual cues. Specifically, graph nodes represent actions, actors, objects, scenes, etc., spatial edges represent spatial (e.g., next to, on top of, etc.) and functional relationships (e.g., attribution, role, etc.) between two nodes with importance weights, and temporal edges represent temporal and causal relationships. We exploit a variety of descriptors in order to capture these rich contextual cues. In the literature, there exist various techniques such as situation recognition \cite{li2017situation}, object detection, scene classification, semantic segmentation, etc. The output of these networks provides embeddings that can serve as the node features of the proposed STGs. 

We perform action segmentation on top of this spatio-temporal graph via stacked spatio-temporal graph convolution. Our STGCN stems from the networks originally proposed for skeleton-based action recognition~\cite{stgcn2018aaai} and introduces two major advancements as our innovations. First, as mentioned before, to accommodate various contextual cues, the nodes of our STG have a wide range of characteristics, leading to the need for using descriptors with varied length. Second, our STG allows arbitrary edge connections (even fully connected graph as an extreme case) to account for the large amount of graph deformation caused by missed detections, occlusions, emerging/disappearing objects, etc. These two advancements are achieved via enhanced designs with additional layers. 

Another innovation we introduce is the use of stacked hourglass STGCN. Stacked hourglass networks using CNNs have been shown to improve results for a number of tasks like human pose estimation~\cite{newell2016stacked}, facial landmark localization~\cite{yang2017stacked}, etc. They allow repeated upsampling and downsampling of features and combine these features at different scales, leading to better performance. We, therefore, propose to apply this encoder-decoder architecture to STGCN. However, different from CNN, STGCN (or more general GCN) employs adjacency matrices to represent irregular connections among nodes. To address this fundamental difference, we adapt the hourglass networks by adding extra steps to down-sample the adjacency matrices at each encoder level to match the compressed dimensions of that level. 

In summary, the proposed Stacked-STGCN offers the following unique innovations: 1) joint inference over a rich set of contextual cues, 2) flexible graph configuration to support a wide range of descriptors with varied feature length and to account for large amounts of graph deformation over long video sequences, and 3) stacked hourglass architecture specifically designed for GCNs including STGCNs. These innovations promise improved recognition/localization accuracy, robustness, and generalization performance for action segmentation over long video sequences. We demonstrate such improvements via our experiments on the CAD120 and Charades datasets. 


\section{Related Works}

\subsection{Neural Networks on Graphs}

In recent years, there have been a number of research directions for applying neural networks on graphs. The original work by Scarselli \etal, referred to as the GNN, was an extension of the recursive neural networks and was used for sub-graph detection\cite{scarselli2005graph}. Later, GNNs were extended and a mapping function was introduced to project a graph and its nodes to an Euclidean space with a fixed dimension \cite{scarselli2009graph}. 
In 2016, Li \etal used gated recurrent units and better optimization techniques to develop the Gated Graph Neural Networks \cite{li2015gated}. 
GNNs have been used in a number of different applications like situation recognition \cite{li2017situation}, human-object interaction \cite{li2018factorizable}, webpage ranking\cite{scarselli2009graph,scarselli2005graph}, mutagenesis\cite{scarselli2009graph}, etc.

The literature also mentions a number of techniques that apply convolutions on graphs. Duvenaud \etal were one of the first to develop convolution operations for graph propagation \cite{duvenaud2015convolutional} whereas Atwood and Towsley developed their own technique independently \cite{atwood2016diffusion}.  Defferrard \etal used approximation in spectral domain \cite{defferrard2016convolutional} based on spectral graph introduced by Hammond \etal \cite{hammond2011wavelets}. In \cite{kipf2016semi}, Kipf and Welling proposed GCNs for semi-supervised classification  based on similar spectral convolutions, but with further simplifications that resulted in higher speed and accuracy. 

\subsection{Action Recognition}

Action recognition is an example of one of the classic computer vision problems being dealt with since the early 1990s. 
In the early days, features like PCA-HOG, SIFT, dense trajectories, etc. were used in conjunction with optimization techniques like HMM, PCA, Markov models, SVM, etc. In 2014, Simonyan and Zisserman used spatial and temporal 2D CNNs \cite{simonyan2014two}. That was followed by the seminal 3D convolutions with combined spatial and temporal convolutional blocks. Since then a series of works following these two schemes, two-stream and 3D convolution, were studied including TSN~\cite{wang2016temporal}, ST-ResNet~\cite{zhang2017deep}, I3D~\cite{carreira2017quo}, P3D~\cite{qiu2017learning}, R(1+2)D~\cite{tran2018closer}, T3D~\cite{diba2017temporal}, S3D~\cite{xiang2018s3d}, etc. 


Another popular type of deep neural networks used for action recognition is the Recurrent Neural Network (RNN) including Long Short-Term Memory networks (LSTM), which are designed to model sequential data. Particularly, RNNs/LSTMs operate on a sequence of per frame features and predict the action label for the whole video sequence (i.e., action recognition) or action of current frame/segment (i.e., action detection/segmentation). The structural-RNN (S-RNN) is one such method that uses RNNs on spatio-temporal graphs for action recognition \cite{jain2016structural}. The S-RNN relies on two independent RNNs, namely nodeRNN and edgeRNN, for iterative spatial and temporal inference. In contrast, our Stacked-STGCN performs joint spatio-temporal inference over a rich set of contextual cues.

Recently, thanks to the rapid development in GNNs, graph-based representation becomes a popular option for action recognition, for instance skeleton-based activity recognition using STGCN~\cite{stgcn2018aaai} and Graph Edge Convolution Networks \cite{zhang2018graph}. In \cite{wang2018videos}, GCN is applied to space-time graphs extracted from the whole video segment to output an accumulative descriptor, which is later combined with the aggregated frame-level features to generate action predictions. Neural Graph Matching Networks were developed for few-shot learning in 3D action recognition \cite{guo2018neural}. 




The most related work is STGCN originally proposed for skeleton-based activity recognition \cite{stgcn2018aaai}. The nodes of the original STGCN are the skeletal joints, spatial connections depend on physical adjacency of these joints in the human body, and temporal edges connect joints of the same type (e.g., right wrist to right wrist) across one consecutive time step.
STGCN on skeleton graph achieves state-of-the-art recognition performance on Kinetics and NTU-RGBD. However, the STG is constructed based on human skeletons, which is indeed an oversimplified structure for the variety and complexity our STG needs to handle in order to perform action segmentation with contextual cues and large graph deformation. Therefore, the original STGCN is not directly applicable. Instead, we use the original STGCN as our basis and introduce a significant amount of augmentation so that STGCN becomes generalizable to a wider variety of applications including action segmentation. 



\subsection{Action Segmentation}
Action segmentation presents a more challenging problem than action recognition in the sense that it requires identifying a sequence of actions with semantic labels and temporally localized starting and ending points of each identified actions. Conditional Random Fields (CRFs) are traditionally used for temporal inference~\cite{mavroudi2018end,pirsiavash2014parsing}. Recently, there has been substantial research interest in leveraging RNNs including LSTM and Gated Recurrent Unit (GRU)~\cite{singh2016multi,yeung2018every}. Lea \etal proposed temporal convolutional networks (TCNs) \cite{rene2017temporal}, which lay the foundation for an additional line of work for action segmentation. Later, a number of variations of TCNs were studied~\cite{ding2017tricornet,ding2018weakly,lei2018temporal}. To the best of our knowledge, no work has attempted to apply STGCNs on a sequence of frame-level scene graph-like representation for action segmentation.  

\begin{figure}
\begin{center}
\fbox{\includegraphics[width=\linewidth]{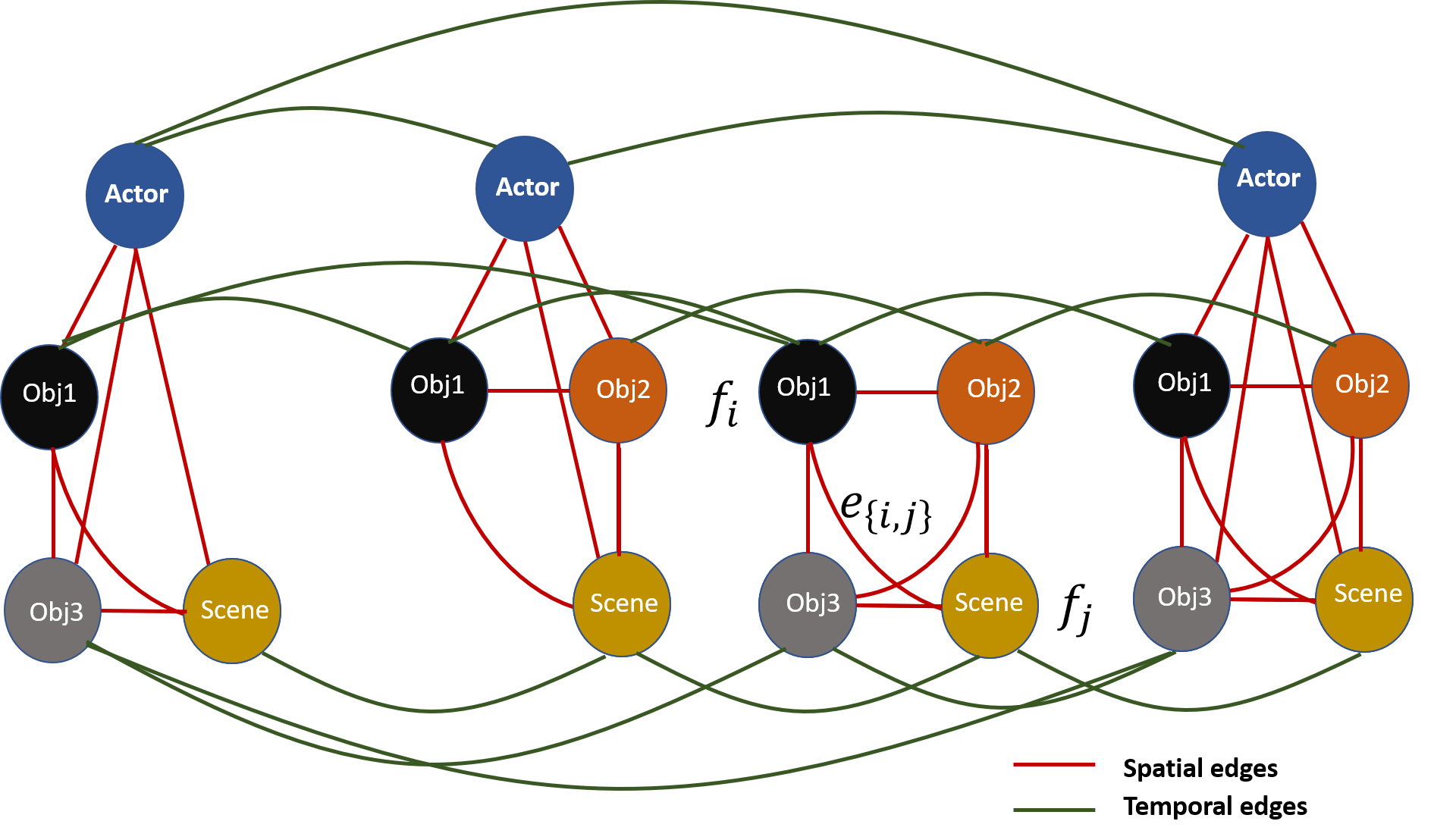}}
\end{center}
   \caption{An illustration of spatio-temporal graphs. Each node $v_i$ is represented by a feature vector denoted by $f_i$. The edge between node $i$ and $j$ has a weight $e_{i,j}$. These edge weights form the spatial and temporal adjacency matrices. Note that our spatio-temporal graph supports a large amount of deformation, such as missed detection (e.g., the actor node and the object 3 node), emerging/disappearing nodes (e.g., the object 2 node), etc.}
\label{fig:simpleGraph}
\end{figure}

\section{Stacked Spatio-Temporal Graph Convolutional Networks}
The proposed Stacked-STGCN is illustrated in Figure \ref{fig:system}. Related notations are given in \ref{sec:gcn}. We describe the basic building block of Stacked-STGCN, i.e., generalized STGCN, in \ref{sec:stgcn} and how to construct the stacked hourglass architecture in \ref{sec:s-stgcn}.

\subsection{Graph Convolutional Networks}
\label{sec:gcn}
Let a graph be defined as $\mathcal{G}(\mathcal{V},\mathcal{E})$ with vertices $\mathcal{V}$ and edges $\mathcal{E}$ (see Figure \ref{fig:simpleGraph}). 
Vertex features of length $d^0$ are denoted as $f_i$  for $i \in \{1,2, \dotsc, N\}$ where $N$ is the total number of nodes.
Edge weights are given as $e_{ij}$ where $e_{ij} \geq 0$ and $i,j \in \{1,2, \dotsc, N\}$.
The graph operation at the \unboldmath$l^{th}$ layer  is defined as:
\begin{align}
\boldsymbol{ H^{l+1} = g(H^{l},A) = \sigma(\hat{D}^{-1/2}\hat{A}\hat{D}^{-1/2}H^{l}W^{l})}
\end{align} 
\boldmath
where $\boldsymbol{W^l}$ and $H^l$ are the $d^{l} \times d^{l+1}$ weight matrix and $N\times d^{l}$ input matrix of the $l^{th}$ layer, respectively. $\hat{A}=I+A$ where $\boldsymbol{A=[e_{i,j}] }$, $\hat{D}$ is the diagonal node degree matrix of $\hat{A}$, and $\sigma$ represents a non-linear activation function (e.g., ReLU). 
\unboldmath


\begin{figure}
\begin{center}
\fbox{\includegraphics[width=\linewidth]{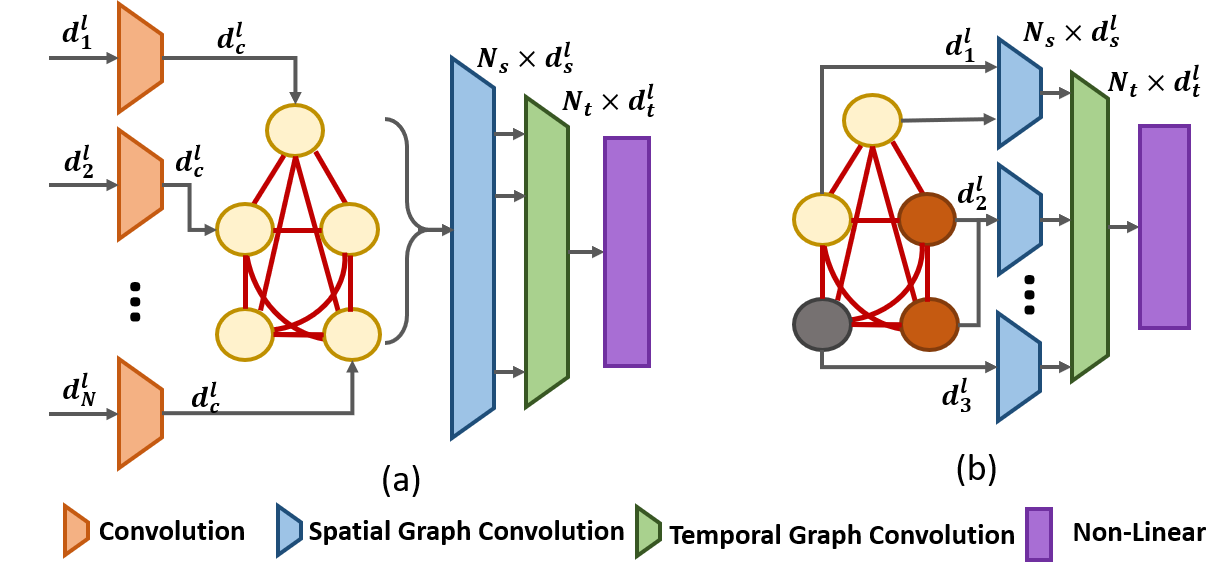}}
\end{center}
\caption{Illustration of two STGCN implementations to support graph nodes with varied feature length. (a) Additional convolution layers to convert node features with varied length to a fixed length. (b) Multiple spatial GCNs each for one cluster of nodes (nodes with the same color) with a similar feature length. These spatial GCNs convert features with varied length to a fixed length.}
\label{fig:STGCN_var}
\end{figure}

\subsection{Spatio-Temporal Graph Convolutional Networks}
\label{sec:stgcn}
STGCN is originally designed for skeleton-based action recognition \cite{stgcn2018aaai}. We apply STGCN for action segmentation of long video sequences using frame-based action graphs extracted via situation recognition \cite{li2017situation}. To accommodate additional application requirements, our STG differs fundamentally in two aspects. First, the original STGCN is based on the human skeletal system with graph nodes corresponding to physical joints and spatial edges representing physical connectivity between these joints. Instead, we use human-object interactions to construct our spatial graph where nodes represent actors, objects, scenes, and actions whereas edges represent their spatial (e.g., next to) and/or functional (e.g., role) relationships. Various descriptors can be extracted either as the channels or nodes of the spatial graph to encode comprehensive contextual information about the actions. For example, we can use pose feature to describe actor nodes, appearance features including attributes at high semantic levels for object nodes, frame-level RGB/flow features for scene nodes, etc.


Second, the original STGCN only connects physical joints of the same type across consecutive time stamps, which indeed reduces to a fixed and grid-like connectivity. As a result, the temporal GCN degrades to conventional convolution. To support  flexible configurations and account for frequent graph deformation in complexity activities (e.g., missed detections,  emerging/disappearing objects, heavy occlusions, etc.), our graph allows arbitrary temporal  connections. For example, an object node present at time $t_0$ can be connected to an object node of the same type at time $t_n$ with $n\geq 1$ in comparison to the original STGNC with $n=1$. 

Let $A_s$ and $A_t$ denote the spatial and temporal adjacency matrices, respectively. Our proposed STGCN operation can be represented mathematically as follows:
\begin{align}
\begin{split}
& \boldsymbol{H^{l+1} = g_t(H^{l}_s,A_{t}) =} \boldsymbol{\sigma(\hat{D_{t}}^{-1/2}\hat{A}_{t}\hat{D_{t}}^{-1/2}H^l_sW^l_t)}\\
& \boldsymbol{H^{l}_s = g_s(H^{l},A_{s}) =}  \boldsymbol{\hat{D_{s}}^{-1/2}\hat{A}_{s}\hat{D_{s}}^{-1/2}H^{l}W^{l}_s}
\end{split}
\end{align}
where $W^l_s$ and $W^l_t$ represents the spatial and temporal weight metrics of the $l^{th}$ convolution layer, respective. In comparison, the original STGCN reduces to
\begin{align}
\begin{split}
& \boldsymbol{H^{l+1} = g(H^{l},A_{s}) =} \\
& \boldsymbol{\sigma(\hat{D_{s}}^{-1/2}\hat{A}_{s}\hat{D_{s}}^{-1/2}H^{l}W^{l}_sW^l_t)}
\end{split}
\end{align}
due to the fixed grid-like temporal connections. 

Note that the original STGCN requires fixed feature length across all graph nodes, which may not hold for our applications where nodes of different types may require different feature vectors to characterize (e.g., features from Situation Recognition are of length 1024 while appearance features from Faster-RCNN\cite{ren2015faster} are of length 2048). To address the problem of varied feature length, one easy solution is to include an additional convolutional layer to convert features with varied length to fixed length (see Figure~\ref{fig:STGCN_var}(a)). However, we argue that nodes of different types may require different length to embed different amounts of information. Converting features to a fixed length may decrease the amount of information they can carry. Therefore, we group nodes into clusters based on their feature length and design multiple spatial GCNs, each corresponding to one of the node cluster. These spatial GCNs will convert features to a fixed length. To allow spatial connections across these node clusters, we model these connections in the temporal adjacency matrix to avoid the use of an additional spatial GCN, since our temporal GCN already allows for arbitrary connections (see Figure~\ref{fig:STGCN_var}(b)).

%
%

Notably, the S-RNN is developed for action recognition in \cite{jain2016structural} where node RNN and edge RNN are used iteratively to process graph-like input. In comparison, our model features a single graph network that can jointly process node features and edge connectivity in an interconnected manner. This, therefore, leads to improved performance and robustness.

\begin{figure*}
\begin{center}
\fbox{\includegraphics[width=0.7\linewidth]{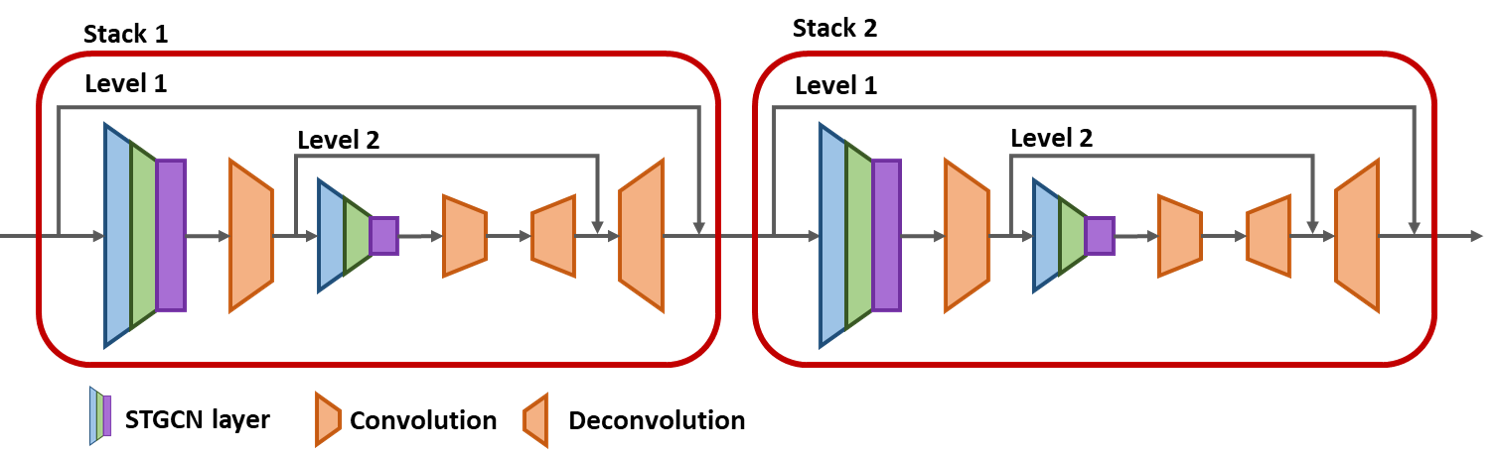}}
\end{center}
   \caption{Illustration of stacked hourglass STGCN with two levels.}
\label{fig:BlockStackSTGCN}
\end{figure*} 

\subsection{Stacking of hourglass STGCN}
\label{sec:s-stgcn}
Hourglass networks consist of a series of downsampling and upsampling operations with skip connections. They follow the principles of the information bottleneck approach to deep learning models \cite{blierdescription} for improved performance. They have also been shown to work well for tasks such as human pose estimation~\cite{newell2016stacked}, facial landmark localization~\cite{yang2017stacked}, etc. In this work, we incorporate the hourglass architecture with STGCN so as to leverage the encoder-decoder structure for  action segmentation with improved accuracy. Particularly, our GCN hourglass network contains a series of a STGCN layer followed by a strided convolution layer as the basic building block for the encoding process. Conventional deconvolution layers comprise the basic unit for the decoding process to bring the spatial and temporal dimensions to the original size. Figure~\ref{fig:BlockStackSTGCN} depicts an example with two levels. 

Note that, at each layer of STGCN, the dimension of the spatial and temporal adjacency matrices, $A_s$ and $A_t$, needs to be adjusted accordingly to reflect the downsampling operation. Take the illustrative example in Figure~\ref{fig:BlockStackSTGCN} for instance and assume that the adjacency matrices $A_{t}$ and $A_{s}$ are of size $N_t\times N_t$ and $N_s\times N_s$, respectively, at level 1 and that a stride of two is used. At level 2, both $A_{t}$ and $A_{s}$ are sub-sampled by two and their dimensions become $N_t/2 \times N_t/2$ and $N_s/2 \times N_s/2$, respectively. Due to the information compression enabled by the encoder-decoder structure, using hourglass networks leads to performance gain compared to using the same number of STGCN layers one after another.

\section{Datasets}
\subsection{CAD120}
The CAD120 dataset is one of the more simplistic datasets available for activity recognition~\cite{koppula2013learning}. It provides RGBD Data for 120 videos on 4 subjects as well as skeletal data. We use the 10 actions classes as our model labels including reaching, moving, pouring, eating, drinking, opening, placing, closing, scrubbing and null. 

The CAD120 dataset splits each video into segments of the above mentioned actions. For each segment, it provides features for object nodes, skeleton features for actor nodes, and spatial weights for object-object and skeleton-object edges. Across segments, it also provides temporal weights for object-object and actor-actor edges. The object node feature captures information about the object's locations in the scene and the way it changes. The Openni's skeleton tracker \cite{openni} is applied to RGBD videos producing skeleton features for actor nodes. The spatial edge weights are based on the relative geometric features among the objects or between an object and the actor. The temporal edge weights capture the changes from one temporal segment to another. Table~\ref{table:CAD120Feature} describes these features in more details.

While experimentation, four fold cross-validation is carried out, where videos from 1 of the 4 people are used for testing and the videos from the rest three for training.

\begin{table}
\begin{center}
\begin{tabular}{|l|}
\hline
\bf{Description} \\
\hline
\hline
\bf{Object Features} \\
N1. Centroid location \\
N2. 2D bounding box \\
N3. Transformation matrix of SIFT matches between \\
adjacent frames \\
N4. Distance moved by the centroid \\
N5. Displacement of centroid \\
\hline
\bf{Actor Features} \\
N6. Location of each joint (8 joints) \\
N7. Distance moved by each joint (8 joints) \\
N8. Displacement of each joint (8 joints) \\
N9. Body pose features\\
N10. Hand position features\\
\hline
\bf{Object-object Features} (computed at start frame,\\
\bf{middle frame, end frame, max and min)}\\
E1. Difference in centroid locations ($\Delta x, \Delta y, \Delta z$) \\
E2. Distance between centroids \\
\hline
\bf{Object–Human Features} (computed at start\\
\bf{frame, middle frame, end frame, max and min)}\\
E3. Distance between each joint location and object\\
centroid\\
\hline
\bf{Object Temporal Features}\\
E4. Total and normalized vertical displacement \\
E5. Total and normalized distance between centroids \\
\hline
\bf{Human Temporal Features}\\
E6. Total and normalized distance between each\\
corresponding joint locations (8 joints) \\
\hline
\end{tabular}
\end{center}
\caption{Features for the CAD120 dataset~\cite{koppula2013learning}.}
\label{table:CAD120Feature}
\end{table}

\subsection{Charades}

\begin{table}
\begin{center}
\begin{tabular}{|l|}
\hline
\bf{Description} \\
\hline
\hline
\bf{Scene Features} \\
N1. FC7 layer output of VGG network trained \\
on RGB frames\\
N2. FC7 layer output of VGG network trained\\
on flow frames\\
N3. I3D pre-final layer output trained on RGB frames \\
N4. I3D pre-final layer output trained on flow frames \\
\hline
\bf{Actor Features} \\
N5.GNN-based Situation Recognition trained \\
on the ImSitu dataset \\
\hline
\bf{Object Features} \\
N6. Top 5 object detection features from Faster-RCNN \\
\hline
\end{tabular}
\end{center}
\caption{Features for the Charades dataset.}
\label{table:CharadesFeature}
\end{table}

The Charades is a recent real-world activity recognition/segmentation dataset including 9848 videos with 157 action classes, 38 object classes, and 33 verb classes~\cite{actorobserver,sigurdsson2016hollywood}. It contains both RGB and flow streams at a frame rate of 24fps. It poses a multi-label, multi-class problem in the sense that at each time step there can be more than one action label. The dataset provides ground-truth object and verb labels as well as FC7 feautres for every $4^{th}$ frames obtained from a two-stream network trained on Charades. The entire dataset is split into 7985 training videos and 1863 testing videos. 

\section{Experiments}


\subsection{CAD120}
We exploited all the node features and edge weights provided by the CAD120 dataset. The skeleton feature of an actor node is of length 630 and the feature of an object node is of length 180. We pass each of these descriptors through convolution layers to convert them to a fixed length of 512. The initial learning rate is 0.0004 and the learning rate scheduler has a drop rate of 0.9 with a step size of 1.

\subsection{Charades}
For the Charades dataset, we explored two types of features, one based on VGG and the other based on I3D \cite{carreira2017quo}, for the scene nodes in our spatio-temporal graph. Further, we used the GNN-based situation recognition technique \cite{li2017situation} trained on the ImSitu dataset \cite{yatskar2016situation} to generate the verb feature for the actor nodes. The top five object features of the Faster-RCNN network trained on MSCOCO are used as descriptors of the object nodes. In total, the spatial dimension of our STG is 8. The VGG features are of length 4096, the verb features 1024, and the object features 2048. Each of these channels are individually processed using convolution layers to convert them to a fixed length (e.g., we used 512). Table~\ref{table:CharadesFeature} summarizes these features. 

In this experiment, spatial nodes are fully connected and temporal edges allow connections across three time steps, i.e., at the $t^{th}$ step there are edges from $t$, to $t+1$ and $t+2$ and $t+3$. The spatial edges between nodes are given a much smaller weight than self connections. We used a stack of three hourglass STGCN blocks. In the model before applying the normalized adjacency matrix, the input is also normalized by subtracting the mean. 
The output of the final Stacked-STGCN block is spatially pooled and passes through a fully connected layer to generate the probability scores of all possible classes. Since the Charades is a multi-label, multi-class dataset, the binary cross-entropy loss was used.
We used an initial learning rate of 0.001 and a learning rate scheduler with a step size of 10 and a drop rate of 0.999.

To further improve action segmentation performance on Charades, we have also used a trained I3D model on Charades to generate descriptors for the scene nodes replacing the VGG features. These feature descriptors are of length 1024. Since I3D already represents short-term temporal dependencies, one block of hourglass STGCN is sufficient for capturing long-term temporal dependencies. The initial learning rate was 0.0005 and the learning rate scheduler was fixed at a drop rate of 0.995 at a step size of 10.


During training, we chose our maximum temporal dimension to be 50. If the length of a video segment is less than 50, we zero-pad the rest of the positions. But these positions are not used for loss or score computation. If the length of a video segment is greater than 50, we randomly select a starting point and use the 50 consecutive frames as the input to our graph.

At test time, we used a sliding window of length 50. Based on overlapping ratios, we applied a weighted average over these windowed scores to produce the final score. We used an overlap of 40 time steps. Following instructions in the Charades dataset, we selected 25 equally spaced points from the available time steps in the video, to generate the final score vectors. 

\section{Results}

\begin{figure*}
\begin{center}
\fbox{\rule{0pt}{2in} \includegraphics[width=\linewidth]{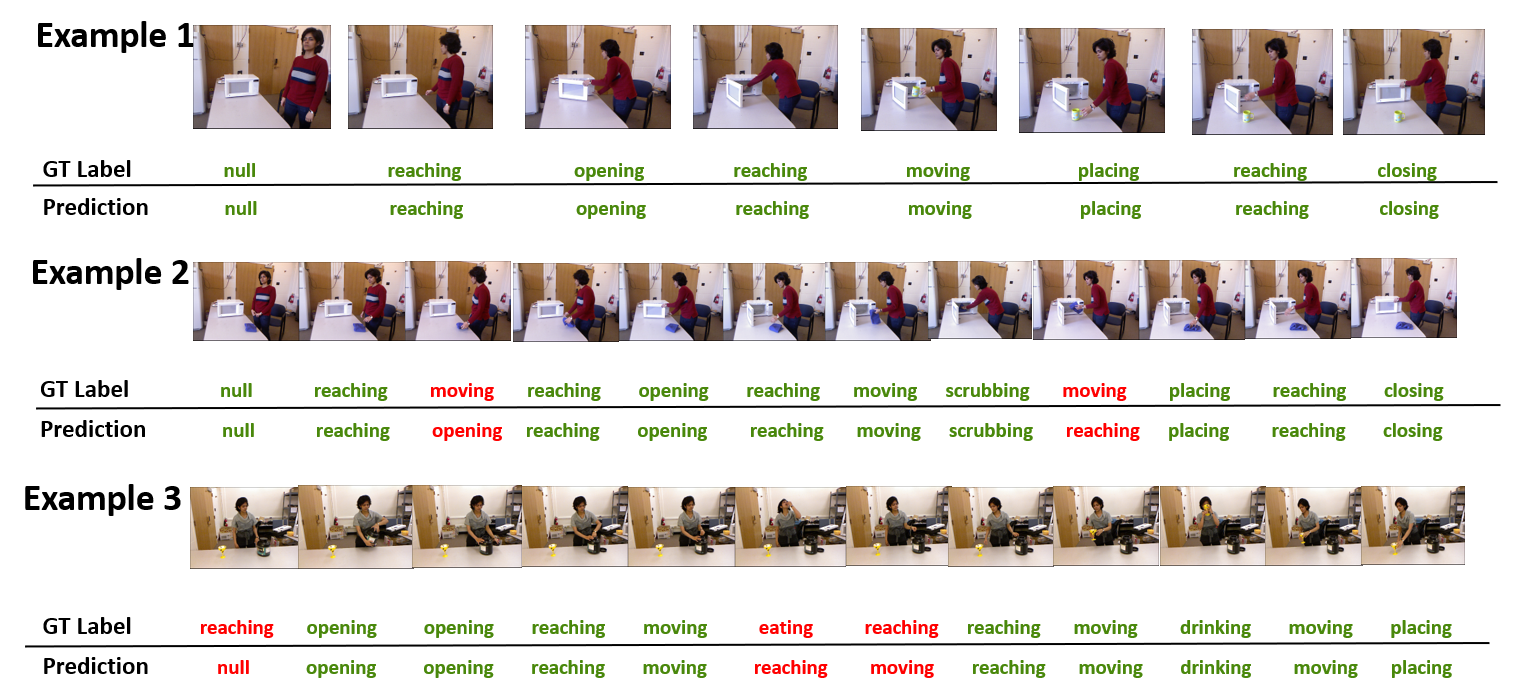}}
\end{center}
   \caption{Action segmentation results of our Stacked-STGCN on the CAD120 dataset. Green: correct detection and red: erroneous detection.}
\label{fig:results_im}
\end{figure*}

\subsection{CAD120}

For the CAD120 dataset, the node features and edge weights are provided by the dataset itself. The same set of features were used by S-RNN \cite{jain2016structural} and Koppula et al \cite{koppula2013learning,koppula2016anticipating} who used spatio-temporal CRF to solve the problem. The S-RNN trains two separate RNN models, one for nodes (i.e., nodeRNN) and the other for edges (i.e., edgeRNN). The edgeRNN is a single layer LSTM of size 128 and the nodeRNN  uses an LSTM of size 256. The actor nodeRNN outputs an action label at each time step. In Table~\ref{table:CAD120_res}, we show some of the previous results, including the best reported one from S-RNN, as well as the result of our STGCN. The F1 score is used as the evaluation metric.  

\begin{table}
\begin{center}
\begin{tabular}{|l|c|}
\hline
\bf{Method} & \bf{F1-score (\%)}\\
\hline\hline
Koppula et al. \cite{koppula2013learning,koppula2016anticipating} & 80.4 \\
S-RNN w/o edge-RNN \cite{jain2016structural}& 82.2 \\
S-RNN \cite{jain2016structural}& 83.2\\
S-RNN(multitask) \cite{jain2016structural} & 82.4\\
\bf{Ours (STGCN)} & \bf{87.21}\\
\hline
\end{tabular}
\end{center}
\caption{Performance comparison based on the F1 score using the CAD120 dataset. Our STGCN improves the F1 score over the best reported result (i.e., S-RNN) by approximately 4.0\%.}
\label{table:CAD120_res}
\end{table}

Our STGCN outperforms the S-RNN by about $4.0\%$ in F1 score. Instead of using two independent RNNs to model interactions among edges and nodes, our STGCN collectively performs joint inference over these inherently interconnected features. This, therefore, leads to the observed performance improvement.

Now looking at Figure~\ref{fig:results_im}, we can see a couple of errors in the second and third examples. For example, the third prediction is \lq opening\rq~ instead of \lq moving\rq~  in the second example. The previous action is \lq reaching\rq~ which is generally what precedes \lq opening\rq~ when the actor is standing in front of a microwave and looking at it. So probably that is the reason for the observed erroneous detection. Also the ninth frame is classified \lq reaching\rq~ instead of \lq moving\rq. If we look at the ninth frame and the eleventh frame, everything appears the same except for the blue cloth in the actor's hand. Our STGCN failed to capture such subtle changes and therefore predicted the wrong action label. 

\subsection{Charades}

\begin{table}
\begin{center}
\begin{tabular}{|l|c|c|c|c|}
\hline
& \bf{Baseline}&\bf{LSTM}&\bf{Super}&\bf{Stacked-STGCN} \\
& & & \bf{-Events} & \bf{(ours)} \\
\hline\hline
VGG  & 6.56 & 7.85 & 8.53 & \bf{10.94} \\
\hline
I3D  & 17.22 & 18.12 & \bf{19.41} & 19.09 \\
\hline
\end{tabular}
\end{center}
\caption{Performance comparison based on mAP between our Stacked-STGCN and the best reported results published in \cite{piergiovanni2018learning} using the Charades dataset.}
\label{table:Charades_res1}
\end{table}

\begin{table}
\begin{center}
\begin{tabular}{|l|c|}
\hline
\bf{Method} & \bf{mAP} \\
\hline\hline
Random \cite{sigurdsson2017asynchronous} & 2.42 \\
RGB \cite{sigurdsson2017asynchronous} & 7.89 \\
Predictive-corrective \cite{dave2017predictivecorrective} & 8.9 \\
Two-Stream \cite{sigurdsson2017asynchronous} & 8.94 \\
Two-Stream + LSTM  \cite{sigurdsson2017asynchronous} & 9.6 \\
R-C3D  \cite{xu2017r} & 12.7 \\
Sigurdsson \etal  \cite{sigurdsson2017asynchronous} & 12.8 \\
I3D \cite{carreira2017quo} & 17.22 \\
I3D +LSTM \cite{piergiovanni2018learning} & 18.1 \\
I3D+Temporal Pyramid \cite{piergiovanni2018learning} & 18.2 \\
I3D + Super-events \cite{piergiovanni2018learning} & \bf{19.41} \\
\hline
VGG + Stacked-STGCN (ours) & 10.94 \\
\hline
VGG+Situation Recognition+ & 11.73\\
FasterRCNN+ Stacked-STGCN (ours) & \\
\hline
I3D +Stacked-STGCN  (ours) & 19.09 \\
\hline
\end{tabular}
\end{center}
\caption{Performance comparison based on mAP with previous works using the Charades dataset.}
\label{table:Charades_res2}
\end{table}

As to the Charades dataset, the mean average precision (mAP) is used as the evaluation metric. For fair comparison, we have used the scripts provided by the Charades dataset to generate mAP scores. 

We examined the performance of Stacked-STGCN using two types of descriptors for the scene nodes, namely frame-based VGG features and segment-based I3D features (see Table~\ref{table:CharadesFeature}). In Table~\ref{table:Charades_res1}, the performance of Stacked-STGCN is compared with a baseline, which uses two-stream VGG or I3D features directly for per frame action label prediction, an LSTM-based method, and the Super-Events approach proposed in \cite{piergiovanni2018learning}. Using VGG features, our Stacked-STGCN yields an approximate $2.4\%$ improvement in mAP. Using I3D features, our Stacked-STGCN ranks the second.  



In Table~\ref{table:Charades_res2}, we compare the performance of Stacked-STGCN against some selected works on Charades. We can see that our Stacked-STGCN outperforms all the methods except for the I3D+super-events \cite{piergiovanni2018learning}, which employs an attention mechanism to learn proper temporal span per class. We believe that incorporating such attention mechanism could further improve the performance of our Stacked-STGCN. For VGG, it improves the best reported result without post-processing \cite{sigurdsson2017asynchronous} by $1.3\%$. 

Another set of results on Charades is from the workshop held in conjunction with CVPR 2017. The results in that competition appear better. However, as mentioned in \cite{piergiovanni2018learning}, that competition used a test set that is different from the validation set we used for performance evaluation. Besides those techniques could have used both the training and validation sets for training. Reference \cite{piergiovanni2018learning} also shows that the same algorithm (i.e., I3D)  that produced 20.72 in the competition produced only 17.22 on the validation set.

\section{Conclusion}

The proposed Stacked-STGCN introduces a stacked hourglass architecture to STGCN for improved generalization performance and localization accuracy. Its building block STGCN is generic enough to take in a variety of nodes/edges and to support flexible graph configuration. In this paper, we applied our Stacked-STGCN to action segmentation and demonstrated improved performances on the CAD120 and Charades datasets. We also note that adding spatial edge connections across nodes with different types lead to only minor performance improvement on Charades. This is mainly due to the oversimplified edge model (i.e., with fixed  weights). Instead of using a binary function to decide on the correlation between these nodes, more sophisticated weights could be explored. We leave this as future work. Finally, we anticipate that thanks to its generic design Stacked-STGCN can be applied to a wider range of applications that require inference over a sequence of graphs with heterogeneous data types and varied temporal extent. 

\section{Acknowledgments}

Supported by the Intelligence Advanced Research Projects Activity (IARPA) via Department of Interior/Interior Business Center (DOI/IBC) contract number D17PC00343. The U.S. Government is authorized to reproduce and distribute reprints for Governmental purposes notwithstanding any copyright annotation thereon. Disclaimer: The views and conclusions contained herein are those of the authors and should not be interpreted as necessarily representing the official policies or endorsements, either expressed or implied, of IARPA, DOI/IBC, or the U.S. Government.

{\small
\bibliographystyle{ieee}
\bibliography{egbib}

\begin{thebibliography}{10}\itemsep=-1pt

\bibitem{openni}
http://openni.org.

\bibitem{atwood2016diffusion}
J.~Atwood and D.~Towsley.
\newblock Diffusion-convolutional neural networks.
\newblock In {\em Advances in Neural Information Processing Systems}, pages
  1993--2001, 2016.

\bibitem{blierdescription}
L.~Blier and Y.~Ollivier.
\newblock The description length of deep learning models.
\newblock In {\em NIPS}, 2018.

\bibitem{bruna2017community}
J.~Bruna and X.~Li.
\newblock Community detection with graph neural networks.
\newblock {\em arXiv preprint arXiv:1705.08415}, 2017.

\bibitem{carreira2017quo}
J.~Carreira and A.~Zisserman.
\newblock Quo vadis, action recognition? a new model and the kinetics dataset.
\newblock In {\em Computer Vision and Pattern Recognition (CVPR), 2017 IEEE
  Conference on}, pages 4724--4733. IEEE, 2017.

\bibitem{dave2017predictivecorrective}
A.~Dave, O.~Russakovsky, and D.~Ramanan.
\newblock Predictivecorrective networks for action detection.
\newblock In {\em Proceedings of the Computer Vision and Pattern Recognition},
  2017.

\bibitem{defferrard2016convolutional}
M.~Defferrard, X.~Bresson, and P.~Vandergheynst.
\newblock Convolutional neural networks on graphs with fast localized spectral
  filtering.
\newblock In {\em Advances in Neural Information Processing Systems}, pages
  3844--3852, 2016.

\bibitem{diba2017temporal}
A.~Diba, M.~Fayyaz, V.~Sharma, A.~H. Karami, M.~M. Arzani, R.~Yousefzadeh, and
  L.~Van~Gool.
\newblock Temporal 3d convnets: New architecture and transfer learning for
  video classification.
\newblock {\em arXiv preprint arXiv:1711.08200}, 2017.

\bibitem{ding2017tricornet}
L.~Ding and C.~Xu.
\newblock Tricornet: A hybrid temporal convolutional and recurrent network for
  video action segmentation.
\newblock {\em arXiv preprint arXiv:1705.07818}, 2017.

\bibitem{ding2018weakly}
L.~Ding and C.~Xu.
\newblock Weakly-supervised action segmentation with iterative soft boundary
  assignment.
\newblock In {\em Proceedings of the IEEE Conference on Computer Vision and
  Pattern Recognition}, pages 6508--6516, 2018.

\bibitem{duvenaud2015convolutional}
D.~K. Duvenaud, D.~Maclaurin, J.~Iparraguirre, R.~Bombarell, T.~Hirzel,
  A.~Aspuru-Guzik, and R.~P. Adams.
\newblock Convolutional networks on graphs for learning molecular fingerprints.
\newblock In {\em Advances in neural information processing systems}, pages
  2224--2232, 2015.

\bibitem{guo2018neural}
M.~Guo, E.~Chou, D.-A. Huang, S.~Song, S.~Yeung, and L.~Fei-Fei.
\newblock Neural graph matching networks for fewshot 3d action recognition.
\newblock In {\em European Conference on Computer Vision}, pages 673--689.
  Springer, 2018.

\bibitem{hammond2011wavelets}
D.~K. Hammond, P.~Vandergheynst, and R.~Gribonval.
\newblock Wavelets on graphs via spectral graph theory.
\newblock {\em Applied and Computational Harmonic Analysis}, 30(2):129--150,
  2011.

\bibitem{jain2016structural}
A.~Jain, A.~R. Zamir, S.~Savarese, and A.~Saxena.
\newblock Structural-rnn: Deep learning on spatio-temporal graphs.
\newblock In {\em Proceedings of the IEEE Conference on Computer Vision and
  Pattern Recognition}, pages 5308--5317, 2016.

\bibitem{kay2017kinetics}
W.~Kay, J.~Carreira, K.~Simonyan, B.~Zhang, C.~Hillier, S.~Vijayanarasimhan,
  F.~Viola, T.~Green, T.~Back, P.~Natsev, et~al.
\newblock The kinetics human action video dataset.
\newblock {\em arXiv preprint arXiv:1705.06950}, 2017.

\bibitem{kipf2016semi}
T.~N. Kipf and M.~Welling.
\newblock Semi-supervised classification with graph convolutional networks.
\newblock {\em arXiv preprint arXiv:1609.02907}, 2016.

\bibitem{koppula2013learning}
H.~S. Koppula, R.~Gupta, and A.~Saxena.
\newblock Learning human activities and object affordances from rgb-d videos.
\newblock {\em The International Journal of Robotics Research}, 32(8):951--970,
  2013.

\bibitem{koppula2016anticipating}
H.~S. Koppula and A.~Saxena.
\newblock Anticipating human activities using object affordances for reactive
  robotic response.
\newblock {\em IEEE transactions on pattern analysis and machine intelligence},
  38(1):14--29, 2016.

\bibitem{lei2018temporal}
P.~Lei and S.~Todorovic.
\newblock Temporal deformable residual networks for action segmentation in
  videos.
\newblock In {\em Proceedings of the IEEE Conference on Computer Vision and
  Pattern Recognition}, pages 6742--6751, 2018.

\bibitem{li2017situation}
R.~Li, M.~Tapaswi, R.~Liao, J.~Jia, R.~Urtasun, and S.~Fidler.
\newblock Situation recognition with graph neural networks.
\newblock {\em arXiv preprint arXiv:1708.04320}, 2017.

\bibitem{li2018factorizable}
Y.~Li, W.~Ouyang, B.~Zhou, J.~Shi, C.~Zhang, and X.~Wang.
\newblock Factorizable net: an efficient subgraph-based framework for scene
  graph generation.
\newblock In {\em European Conference on Computer Vision}, pages 346--363.
  Springer, 2018.

\bibitem{li2015gated}
Y.~Li, D.~Tarlow, M.~Brockschmidt, and R.~Zemel.
\newblock Gated graph sequence neural networks.
\newblock {\em arXiv preprint arXiv:1511.05493}, 2015.

\bibitem{mavroudi2018end}
E.~Mavroudi, D.~Bhaskara, S.~Sefati, H.~Ali, and R.~Vidal.
\newblock End-to-end fine-grained action segmentation and recognition using
  conditional random field models and discriminative sparse coding.
\newblock {\em arXiv preprint arXiv:1801.09571}, 2018.

\bibitem{newell2016stacked}
A.~Newell, K.~Yang, and J.~Deng.
\newblock Stacked hourglass networks for human pose estimation.
\newblock In {\em European Conference on Computer Vision}, pages 483--499.
  Springer, 2016.

\bibitem{piergiovanni2018learning}
A.~Piergiovanni and M.~S. Ryoo.
\newblock Learning latent super-events to detect multiple activities in videos.
\newblock In {\em Proceedings of the IEEE Conference on Computer Vision and
  Pattern Recognition (CVPR)}, volume~4, 2018.

\bibitem{pirsiavash2014parsing}
H.~Pirsiavash and D.~Ramanan.
\newblock Parsing videos of actions with segmental grammars.
\newblock In {\em Proceedings of the IEEE Conference on Computer Vision and
  Pattern Recognition}, pages 612--619, 2014.

\bibitem{qiu2017learning}
Z.~Qiu, T.~Yao, and T.~Mei.
\newblock Learning spatio-temporal representation with pseudo-3d residual
  networks.
\newblock In {\em 2017 IEEE International Conference on Computer Vision
  (ICCV)}, pages 5534--5542. IEEE, 2017.

\bibitem{ren2015faster}
S.~Ren, K.~He, R.~Girshick, and J.~Sun.
\newblock Faster r-cnn: Towards real-time object detection with region proposal
  networks.
\newblock In {\em Advances in neural information processing systems}, pages
  91--99, 2015.

\bibitem{rene2017temporal}
C.~L. M. D.~F. Ren{\'e} and V.~A. R. G.~D. Hager.
\newblock Temporal convolutional networks for action segmentation and
  detection.
\newblock In {\em IEEE International Conference on Computer Vision (ICCV)},
  2017.

\bibitem{scarselli2009graph}
F.~Scarselli, M.~Gori, A.~C. Tsoi, M.~Hagenbuchner, and G.~Monfardini.
\newblock The graph neural network model.
\newblock {\em IEEE Transactions on Neural Networks}, 20(1):61--80, 2009.

\bibitem{scarselli2005graph}
F.~Scarselli, S.~L. Yong, M.~Gori, M.~Hagenbuchner, A.~C. Tsoi, and M.~Maggini.
\newblock Graph neural networks for ranking web pages.
\newblock In {\em Proceedings of the 2005 IEEE/WIC/ACM International Conference
  on Web Intelligence}, pages 666--672. IEEE Computer Society, 2005.

\bibitem{schlichtkrull2018modeling}
M.~Schlichtkrull, T.~N. Kipf, P.~Bloem, R.~van~den Berg, I.~Titov, and
  M.~Welling.
\newblock Modeling relational data with graph convolutional networks.
\newblock In {\em European Semantic Web Conference}, pages 593--607. Springer,
  2018.

\bibitem{shahroudy2016ntu}
A.~Shahroudy, J.~Liu, T.-T. Ng, and G.~Wang.
\newblock Ntu rgb+ d: A large scale dataset for 3d human activity analysis.
\newblock In {\em Proceedings of the IEEE conference on computer vision and
  pattern recognition}, pages 1010--1019, 2016.

\bibitem{sigurdsson2017asynchronous}
G.~A. Sigurdsson, S.~K. Divvala, A.~Farhadi, and A.~Gupta.
\newblock Asynchronous temporal fields for action recognition.
\newblock In {\em CVPR}, volume~5, page~7, 2017.

\bibitem{actorobserver}
G.~A. Sigurdsson, A.~Gupta, C.~Schmid, A.~Farhadi, and K.~Alahari.
\newblock Actor and observer: Joint modeling of first and third-person videos.
\newblock In {\em CVPR}, 2018.

\bibitem{sigurdsson2016hollywood}
G.~A. Sigurdsson, G.~Varol, X.~Wang, A.~Farhadi, I.~Laptev, and A.~Gupta.
\newblock Hollywood in homes: Crowdsourcing data collection for activity
  understanding.
\newblock In {\em European Conference on Computer Vision}, 2016.

\bibitem{simonyan2014two}
K.~Simonyan and A.~Zisserman.
\newblock Two-stream convolutional networks for action recognition in videos.
\newblock In {\em Advances in neural information processing systems}, pages
  568--576, 2014.

\bibitem{simonyan2014very}
K.~Simonyan and A.~Zisserman.
\newblock Very deep convolutional networks for large-scale image recognition.
\newblock {\em arXiv preprint arXiv:1409.1556}, 2014.

\bibitem{singh2016multi}
B.~Singh, T.~K. Marks, M.~Jones, O.~Tuzel, and M.~Shao.
\newblock A multi-stream bi-directional recurrent neural network for
  fine-grained action detection.
\newblock In {\em Proceedings of the IEEE Conference on Computer Vision and
  Pattern Recognition}, pages 1961--1970, 2016.

\bibitem{te2018rgcnn}
G.~Te, W.~Hu, Z.~Guo, and A.~Zheng.
\newblock Rgcnn: Regularized graph cnn for point cloud segmentation.
\newblock {\em arXiv preprint arXiv:1806.02952}, 2018.

\bibitem{tran2018closer}
D.~Tran, H.~Wang, L.~Torresani, J.~Ray, Y.~LeCun, and M.~Paluri.
\newblock A closer look at spatiotemporal convolutions for action recognition.
\newblock In {\em Proceedings of the IEEE Conference on Computer Vision and
  Pattern Recognition}, pages 6450--6459, 2018.

\bibitem{wang2016temporal}
L.~Wang, Y.~Xiong, Z.~Wang, Y.~Qiao, D.~Lin, X.~Tang, and L.~Van~Gool.
\newblock Temporal segment networks: Towards good practices for deep action
  recognition.
\newblock In {\em European Conference on Computer Vision}, pages 20--36.
  Springer, 2016.

\bibitem{wang2018videos}
X.~Wang and A.~Gupta.
\newblock Videos as space-time region graphs.
\newblock {\em arXiv preprint arXiv:1806.01810}, 2018.

\bibitem{xiang2018s3d}
X.~Xiang, Y.~Tian, A.~Reiter, G.~D. Hager, and T.~D. Tran.
\newblock S3d: Stacking segmental p3d for action quality assessment.
\newblock In {\em 2018 25th IEEE International Conference on Image Processing
  (ICIP)}, pages 928--932. IEEE, 2018.

\bibitem{xu2017r}
H.~Xu, A.~Das, and K.~Saenko.
\newblock R-c3d: region convolutional 3d network for temporal activity
  detection.
\newblock In {\em IEEE Int. Conf. on Computer Vision (ICCV)}, pages 5794--5803,
  2017.

\bibitem{stgcn2018aaai}
S.~Yan, Y.~Xiong, and D.~Lin.
\newblock Spatial temporal graph convolutional networks for skeleton-based
  action recognition.
\newblock In {\em AAAI}, 2018.

\bibitem{yang2017stacked}
J.~Yang, Q.~Liu, and K.~Zhang.
\newblock Stacked hourglass network for robust facial landmark localisation.
\newblock In {\em Computer Vision and Pattern Recognition Workshops (CVPRW),
  2017 IEEE Conference on}, pages 2025--2033. IEEE, 2017.

\bibitem{yatskar2016situation}
M.~Yatskar, L.~Zettlemoyer, and A.~Farhadi.
\newblock Situation recognition: Visual semantic role labeling for image
  understanding.
\newblock In {\em Proceedings of the IEEE Conference on Computer Vision and
  Pattern Recognition}, pages 5534--5542, 2016.

\bibitem{yeung2018every}
S.~Yeung, O.~Russakovsky, N.~Jin, M.~Andriluka, G.~Mori, and L.~Fei-Fei.
\newblock Every moment counts: Dense detailed labeling of actions in complex
  videos.
\newblock {\em International Journal of Computer Vision}, 126(2-4):375--389,
  2018.

\bibitem{zhang2017deep}
J.~Zhang, Y.~Zheng, and D.~Qi.
\newblock Deep spatio-temporal residual networks for citywide crowd flows
  prediction.
\newblock In {\em AAAI}, pages 1655--1661, 2017.

\bibitem{zhang2018graph}
X.~Zhang, C.~Xu, and D.~Tao.
\newblock Graph edge convolutional neural networks for skeleton based action
  recognition.
\newblock {\em arXiv preprint arXiv:1805.06184}, 2018.

\end{thebibliography}
}

\end{document}